\documentclass[sigconf]{acmart}
\AtBeginDocument{%
  }

\usepackage{multirow} 
\usepackage{colortbl}  

\acmSubmissionID{3241}

\copyrightyear{2025}
\acmYear{2025}
\setcopyright{acmlicensed}\acmConference[MM '25]{Proceedings of the 33rd ACM International Conference on Multimedia}{October 27--31, 2025}{Dublin, Ireland}
\acmBooktitle{Proceedings of the 33rd ACM International Conference on Multimedia (MM '25), October 27--31, 2025, Dublin, Ireland}
\acmDOI{10.1145/3746027.3755328}
\acmISBN{979-8-4007-2035-2/2025/10}

\begin{document}

\title{Dynamic 2D Gaussians: Geometrically Accurate Radiance Fields
for Dynamic Objects}

\author{Shuai Zhang}
\authornote{Both authors contributed equally to this research.}
\affiliation{%
  \department{School of EIC}
  \institution{Huazhong University of Science and Technology}
  \city{Wuhan}
  \state{Hubei}
  \country{China}
}
\email{shuaizhang@hust.edu.cn}

\author{Guanjun Wu}
\authornotemark[1]
\affiliation{%
  \department{School of CS}
  \institution{Huazhong University of Science and Technology}
  \city{Wuhan}
  \state{Hubei}
  \country{China}}
\email{guajuwu@hust.edu.cn}

\author{Zhoufeng Xie}
\affiliation{%
  \department{School of EIC}
  \institution{Huazhong University of Science and Technology}
  \city{Wuhan}
  \state{Hubei}
  \country{China}
}
\email{162130328@nuaa.edu.cn}

\author{Xinggang Wang}
\affiliation{%
 \department{School of EIC}
 \institution{Huazhong University of Science and Technology}
 \city{Wuhan}
 \state{Hubei}
 \country{China}}
\email{xgwang@hust.edu.cn}

\author{Bin Feng}
\affiliation{%
 \department{School of EIC}
  \institution{Huazhong University of Science and Technology}
  \city{Wuhan}
  \state{Hubei}
  \country{China}}
\email{fengbin@hust.edu.cn}

\author{Wenyu Liu}
\authornote{Wenyu Liu is the corresponding author.}
\affiliation{%
 \department{School of EIC}
  \institution{Huazhong University of Science and Technology}
  \city{Wuhan}
  \state{Hubei}
  \country{China}}
\email{liuwy@hust.edu.cn}

\renewcommand{\shortauthors}{Shuai Zhang, et al.}

\begin{abstract}
Reconstructing objects and extracting high-quality surfaces play a vital role in the real world. Current 4D representations show the ability to render high-quality novel views for dynamic objects, but cannot reconstruct high-quality meshes due to their implicit or geometrically inaccurate representations. In this paper, we propose a novel representation that can reconstruct accurate meshes from sparse image input, named Dynamic 2D Gaussians (D-2DGS). We adopt 2D Gaussians for basic geometry representation and use sparse-controlled points to capture the 2D Gaussian's deformation. By extracting the object mask from the rendered high-quality image and masking the rendered depth map, we remove floaters that are prone to occur during reconstruction and can extract high-quality dynamic mesh sequences of dynamic objects. Experiments demonstrate that our D-2DGS is outstanding in reconstructing detailed and smooth high-quality meshes from sparse inputs. The code is available at https://github.com/hustvl/Dynamic-2DGS.
\end{abstract}
\vspace{-20pt}



\vspace{-10pt}
\vspace{-10pt}
\begin{teaserfigure}
\centering
  \includegraphics[width=0.84\textwidth]{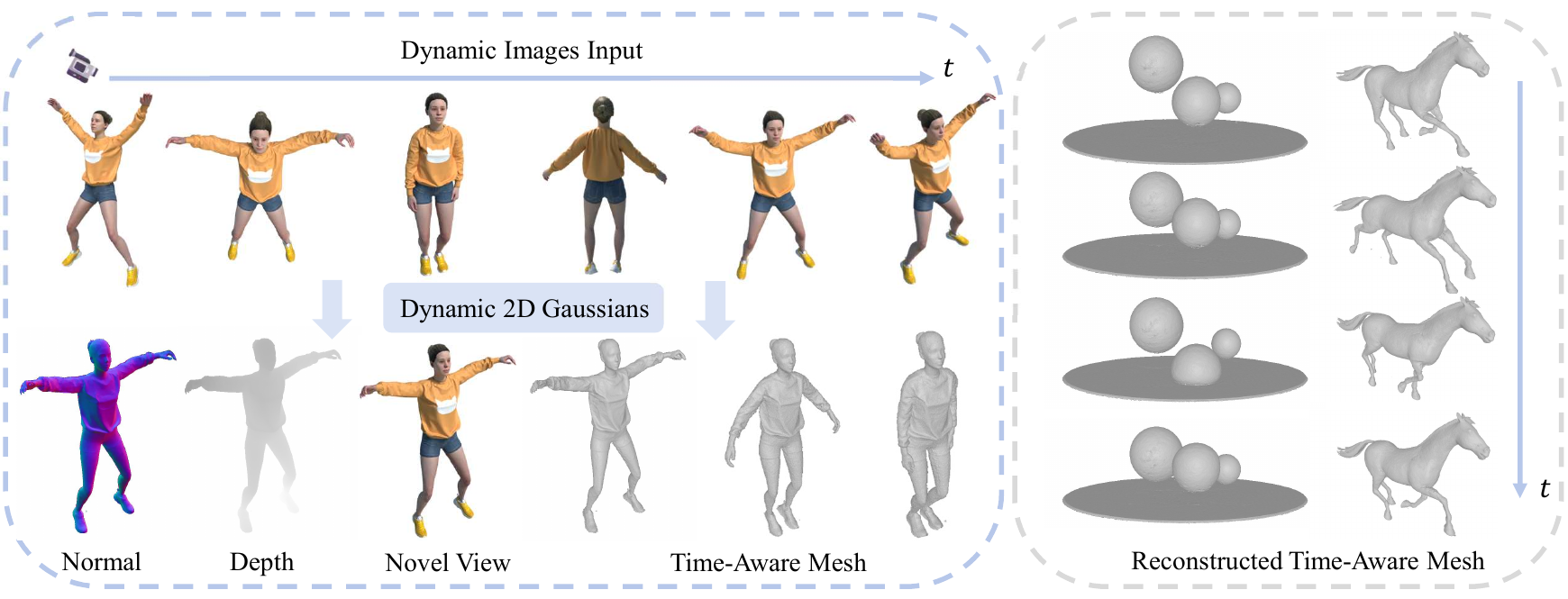}
  \vspace{-5pt}
  \caption{Applications of D-2DGS. Given sparse 2D dynamic images, D-2DGS can reconstruct dynamic objects and extract high-quality normal, time-aware mesh, render quality depth, and novel view images. }
  \label{fig:teaser}
\end{teaserfigure}
\vspace{-20pt}

\maketitle

\section{Introduction}
The natural world is dynamic, and how to extract the geometric shape of dynamic objects from the sparse input images is an important research problem in computer vision. The main challenges lie in the complex and non-rigid motion, which makes it hard to model motion accurately.
Some 3D representations ~\cite{wang2023neus2,huang20242d} succeed in reconstructing high-quality mesh from static objects but may fall short in 4D objects since we cannot leave every life stock still. 

Many approaches succeed in representing dynamic scenes and rendering high-quality novel views. Dynamic NeRFs~\cite{pumarola2021dnerf,park2021nerfies,ParkSHBBGMS21} mainly suffer from their implicit representations, which cause unfriendly memory consumption. Dynamic Gaussian splatting~\cite{wu20244d,yang2024deformable,huang2024sc} representations can maintain a set of explicit 3D Gaussians~\cite{3dgs} for efficient NVS with high training efficiency, and the proposed deformation field shows the potential to model an object's motion accurately. However, 3D Gaussians are mainly for NVS and do not have multiview consistency, which causes inaccurate geometry~\cite{huang20242d}. Some 3D Gaussian-based approaches also succeed in reconstructing a 3D mesh from dynamic images. We also found it~\cite{Liu0025a} lacks high-quality details. We believe that a proper representation that can reconstruct dynamic objects may achieve two goals: (a) modeling complex motion with high efficiency and (b) enabling the export of high-quality and smooth geometry.

Recently, 2D Gaussian Splatting (2DGS)~\cite{huang20242d} uses a collection of two-dimensional oriented planar Gaussian disks to represent scenes, ensuring consistent view geometry when modeling surfaces. Extending 2D Gaussians to dynamic object reconstruction and get a high-quality surface is a natural but difficult topic. There are two challenges, and the first problem is \textit{How to extract a clean, consistent surface from sparse input?} \textit{Geometry floaters} also exist in the deformation of 2D Gaussian primitives: the Gaussian owns the same color as the background, which is hard to be pruned and would induce inaccurate depth, triggering the low-quality mesh extracted by Truncated Signed Distance Function (TSDF). Moreover, \textit{modeling an accurate object's motion} also remains a hard issue in the dynamic representations. It's hard to preserve correspondence between 2D Gaussians' motion and real-world deformation, which may lead to unstable surfaces. Meanwhile, We observe that while 2D Gaussians benefit from multiview consistency due to their isotropic characteristic, it also degrades the Gaussian's fitting ability when dealing with sparse views.

 To tackle the aforementioned problems, we propose to use the sparse-controlled points method~\cite{huang2024sc} to model 2D Gaussians' motion. Our findings indicate that sparse-controlled points accurately model semi-rigid motion because nearby 2D Gaussian surfels are determined by the sparse-controlled points, which provide a more precise canonical-world transformation, ensuring the geometric accuracy of 2D Gaussians in the canonical space and resulting in smoother surfaces. Furthermore, we introduce a filtering method to remove \textit{geometry floaters} by filtering the depth image using the rendered high-quality RGB mask. This approach allows for an accurate surface mesh to be obtained via the TSDF method.

Our contributions can be summarized as follows:

\vspace{-10pt}
\begin{itemize}
    \item We propose Dynamic 2D Gaussians (D-2DGS), a novel framework that employs sparse-controlled points to guide 2D Gaussians' deformation and reconstruct the accurate dynamic mesh. Our framework takes into account both the dynamic and geometric properties of objects.
    \item To extract a clean mesh, it is necessary to remove the \textit{Geometry floaters} generated during reconstruction. Therefore, we propose to filter the depth image using a mask extracted from the rendered high-quality RGB image.
    \item Experiments demonstrate that our D-2DGS achieves state-of-the-art (SOTA) reconstruction/mesh rendering quality compared with other advanced representations.
\end{itemize}

\begin{figure}[t]
\centering
\includegraphics[width=0.45\textwidth]{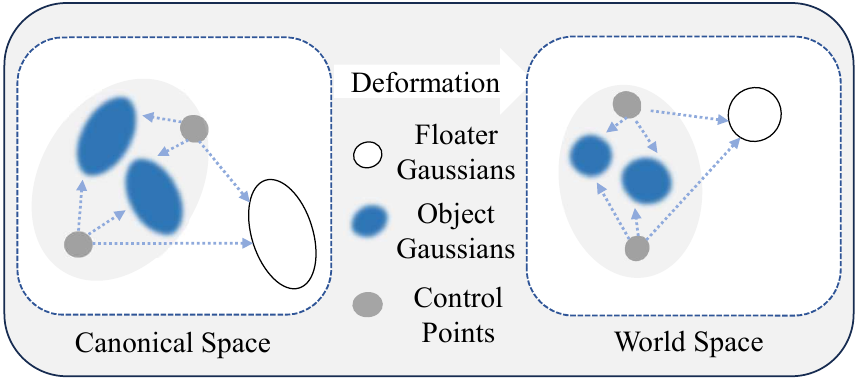}
\caption{Illustration of \textit{Geometry floaters}. After deformation, both objects Gaussians and floater Gaussians are transformed by control points. The similar color to the background makes it hard to be pruned, which causes inaccurate geometry.}
\label{fig:motivation1}
\vspace{-15pt}
\end{figure}

\section{Related Works}
\subsection{3D/4D Representations} 
\label{subsec:3D/4D representations}
Representing the 4D world and rendering photo-realistic novel view images has been an important and challenging topic. Explicit representations such as mesh~\cite{li20184d,guo2015robust}, voxels~\cite{guo2019relightables,hu2022hvtr,li2017robust}, and point clouds~\cite{pointnerf} enjoy editable manipulations and GPU-friendly applications but suffer from strict multi-view constraints or more sensors. Recently, implicit~\cite{martin2021nerf,park2021nerfies,ParkSHBBGMS21} representations have demonstrated their efficiency for novel view synthesis from sparse input, and many approaches~\cite{kplanes,hexplane,tensor4d,fang2022fast} combine the benefits of these representations. Notably, Gaussian Splatting~\cite{3dgs,yang2024realtime,wu20244d,yang2024deformable,zhang2024togs,lei2025mosca} (GS) based representations as an efficient mixture presentation, show the ability to render high-quality novel views and support many downstream tasks~\cite{jiang2024vr,ji2024sa4d,yi2024gaussiandreamer, yi2024gaussiandreamerpro}. Recent 4D representation, SC-GS~\cite{huang2024sc}, proposes a sparse control points method to control the deformation of 3D Gaussians, which reaches outstanding NVS quality and high training efficiency. However, most 3D/4D representations are only for NVS or extracting mesh from static scenes, and there is less research focus on extracting an accurate mesh from dynamic objects. Our D-2DGS shows the ability to reconstruct an accurate 3D mesh from 2D images captured in the 4D world.

\subsection{Geometry Reconstruction}
\label{subsec: Geometry Reconstruction}

Geometry reconstructions are mainly from streamable multiview constraints~\cite{collet2015high,li20184d,guo2015robust}. Recently, 3DGS~\cite{3dgs} is proven to enjoy high training efficiency and shows great potential to reconstruct high-quality mesh from spatial Gaussians.
Many works are proposed to improve the geometric accuracy of 3DGS. SuGaR~\cite{guedon2024sugar} introduces a regularization term that encourages the Gaussians to align well with the surface of the scene. NeuSG~\cite{chen2023neusg} jointly optimizes 3DGS and neural implicit models, combining the advantages of both, enabling the generation of complete surfaces with high details.
Similarly, 3DGSR~\cite{lyu20243dgsr} also encourages the alignment of 3D Gaussian and SDF to improve the surface reconstruction effect. 2D Gaussian Splatting (2DGS)~\cite{huang20242d}is a breakthrough in mesh reconstruction since the view geometry consistency is maintained by planar Gaussian disks. Recently, 3D foundation models ~\cite{xiang2025structured,tochilkin2024triposr,li2025triposg,ye2025hi3dgen} enable high-quality mesh generation from sparse input.  

The above geometric reconstruction methods are all for static scenes. Recently, some works have been attempted on the geometric reconstruction of dynamic scenes. Dynamic Gaussian Mesh (DGMesh)~\cite{Liu0025a} aims to reconstruct a high-fidelity and temporally consistent mesh from a given monocular video. DGMesh introduces Gaussian grid anchoring to encourage uniformly distributed Gaussian distributions and obtains better grid reconstruction by grid densification and pruning of deformed Gaussian distributions. MaGS~\cite{ma2024reconstructing} proposes a mutual-adsorbed mesh Gaussian representation to refine the generated mesh. Vidu4D~\cite{0001WCW0024} uses Dynamic Gaussian Surfels as scene representation primitives, achieving text-to-4D generation with high-quality appearance and geometry. Here are many concurrent works~\cite{ma2024reconstructing,li2024dgns} that propose to use a 3D Gaussian representation to reconstruct a mesh from dynamic scenes. Papers\cite{cai2024dynasurfgs,wang2024space} adopt a deformation field network and Gaussian surfels to model the scene's deformation. Our methods propose to model Gaussian surfels' motion by sparse-controlled points, maintaining consistent and smooth surface mesh generation.

\section{Preliminaries}

\subsection{2D Gaussian Splatting} 
2D Gaussian Splatting (2DGS) simplifies 3D modeling through 2D Gaussians. Each 2D Gaussians is represented by a central point $P_{k}$, two principal tangential vectors $\mathbf{t}_{u}$ and $\mathbf{t}_{v}$, and a scaling vector $\mathbf{S}=(s_{u},s_{v})$ that controls the variance of the 2D Gaussian. The original normal of the 2D Gaussian is defined by two orthogonal tangent vectors $\mathbf{t}_{w}=\mathbf{t}_{u}\times \mathbf{t}_{v}$. The 2D Gaussian function can be parameterized on the local tangent plane in world space as:

\begin{equation}
P(u,v)=P_{k}+s_{u}\mathbf{t} _{u}u+s_{v}\mathbf{t} _{v}v.
\end{equation}

For a point $\mathbf{u}=(u,v)$ in the $uv$ space, its 2D Gaussian value is:

\begin{equation}
\mathcal{G} (\mathbf{u})=exp(-\frac{u^{2}+v^{2}}{2}).
\end{equation}

The parameters of each 2D Gaussian also include opacity $\alpha$ and view-dependent appearance $c$ parameterized with spherical harmonics.

2D Gaussians are sorted by their center depth and organized into tiles based on their bounding boxes. The alpha-weighted appearance is integrated from front to back using volumetric alpha blending:

\begin{equation}
c(x)=\sum_{i=1}\mathbf{c} _{i}\alpha \mathcal{G} (\mathbf{u} (x))\prod_{j=1}^{i-1}(1-\alpha _{j}\mathcal{G}(\mathbf{u} (x))) .
\end{equation}

However, 2D Gaussians can only reconstruct a 3D mesh from static scenes. We propose to use Sparse-Controlled points to model 2D Gaussians deformation and extend 2D Gaussians to represent the 4D world successfully.

\begin{figure*}[!h]
\centering

\includegraphics[width=1\textwidth]{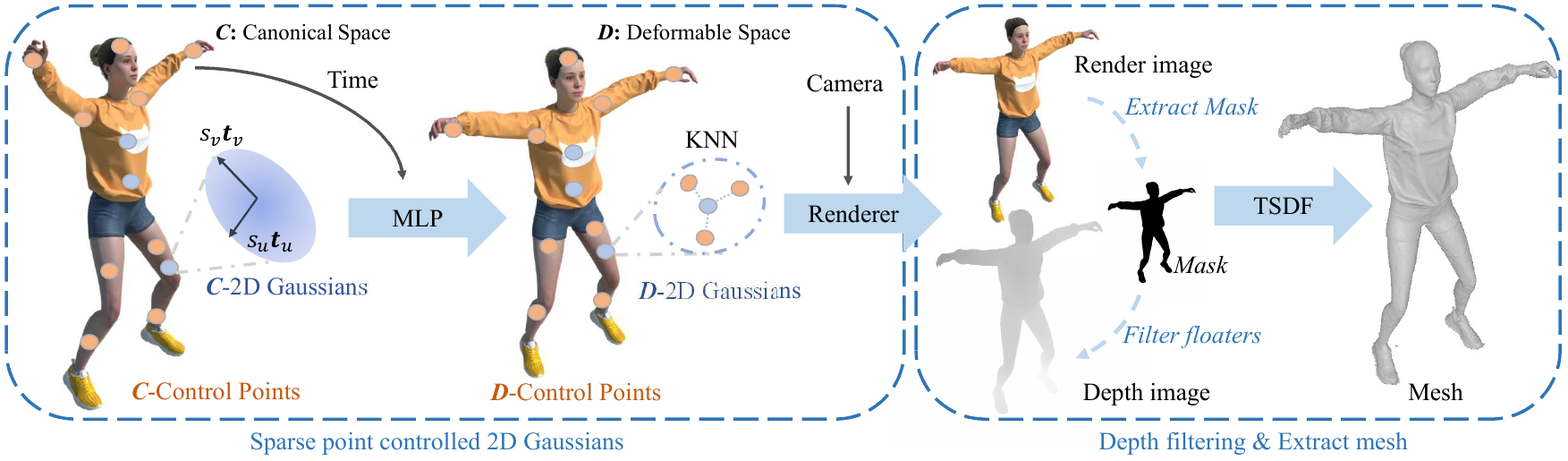}
\vspace{-15pt}
\caption{Framework of our D-2DGS. Sparse points are bonded with canonical 2D Gaussians. Deformation networks are used to predict each sparse control point's control signals given any timestamp. The image and depth are rendered by deformed 2D Gaussians with alpha blending. To get high-quality meshes, depth images are filtered by rendered images with RGB mask, and then TSDF is applied on multiview depth images and RGB images.}
\vspace{-10pt}
\label{fig2}
\end{figure*}

\subsection{Sparse-Controlled Gaussian Splatting} 
Sparse-Controlled Gaussian Splatting (SC-GS)~\cite{huang2024sc} shows excellent performance in the task of synthesizing novel views in dynamic scenes. It uses sparse control points to control dense 3D Gaussians in the scene. For each control point at timestep $t$, SC-GS uses MLP $\Phi$ to predict a rotation matrix $R_{i}^{t}\in \mathbf{SE} (3)$ and translation matrix $T_{i}^{t}\in \mathbb{R} ^{3}$. For each Gaussian $G_{j}:(\mu_{j},q_{j},s_{j},\sigma _{j}, sh_{j} )$, use k-nearest neighbor (KNN) search to get the $K$ neighboring control points $\{p_{i},i=0,1,...,k\}$. SC-GS employs LBS~\cite{sumner2007embedded} $\mathcal{I}$ to compute the warped Gaussian $\mu _{j}^{t}$ and $q_{j}^{t}$:

\begin{equation}
\mu _{j}^{t}=\sum_{k\in \mathcal{N} _{j}}w_{jk}(R_{k}^{t}(\mu _{j}-p_{k}))+p_{k}+T_{k}^{t}),
\label{eq:lbs1}
\end{equation}

\begin{equation}
q_{j}^{t}=(\sum_{k\in \mathcal{N} _{j}}w_{jk}r_{k}^{t} )\otimes q_{j},
\label{eq:lbs2}
\end{equation}

\noindent
where $R_{k}^{t}$ and $r_{k}^{t}$ are the matrix and quaternion representations of the predicted rotation at control point $k$, respectively. The weight $w_{jk}$ is calculated by:

\begin{equation}
w_{jk}=\frac{\hat{w}_{jk} }{\sum_{k\in \mathcal{N} _{j}}\hat{w}_{jk}  }  ,\mathrm{where} ~\hat{w}_{jk}=exp(-\frac{d_{jk}^{2}}{2o_{k}^{2}} ),
\end{equation}

\noindent
where $d_{jk}$ is the distance between the Gaussian and the adjacent control point, and $o_{k}$ is the learned radius parameter of the control point.

However, SC-GS~\cite{huang2024sc} mainly chooses 3D-GS~\cite{3dgs} as its geometry representation, leading to multiview inconsistency. Meanwhile, there are fewer geometry constraints during the optimization of canonical Gaussians, we mainly choose 2DGS~\cite{huang20242d} as basic geometry, and propose an outlier removal method to overcome the \textit{Geometry floaters} as shown in Fig.~\ref{fig:motivation1}.

\section{Method}

\subsection{Dynamic 2D Gaussians Framework}
The pipeline of our framework is illustrated in Fig.~\ref{fig2}. We adopt 2D Gaussian primitives $\mathcal{G}$ for scene representation. We follow deformation-based Gaussian splatting to compute the deformation of each 2D Gaussian at timestamp $t_j$:

\begin{equation}
\mathcal{G}^j = \Phi(\mathcal{G},t_j,\mathbf{P}),
\end{equation}

\noindent
where $\Phi$ denotes deformation network, $\mathbf{P}=\{p_i\}_{i=1}^N$ denotes sparse control points. Then $\mathcal{G}^j$ are used to render the RGB image $\mathbf{I}$ and the depth image $\mathbf{D}$ by differential splatting algorithm~\cite{huang2024sc,yifan2019differentiablesplatting} $\mathcal{S}$ with a camera parameters $[\mathbf{R},\mathbf{t}]$:

\begin{equation}
    \mathbf{I}, \mathbf{D} = \mathcal{S}\left((\mathcal{G}^j; \theta), [\mathbf{R},\mathbf{t}] \right).
\end{equation}

We then extract a mask $\mathbf{M}$ from the rendered image $\mathbf{I}$ and apply it to the depth image $\mathbf{D}$ to filter out floaters:

\begin{equation}
\label{mask flitering}
    \mathbf{D}' = Flitering(\mathbf{I},\mathbf{D}),
\end{equation}
the $Flitering$ progress will be discussed later.

Finally, the refined depth image $\mathbf{D}'$ is used to extract the mesh using the Truncated Signed Distance Function (TSDF).

\subsection{Sparse point controlled 2D Gaussians}

To accurately extract dynamic mesh sequences, it is essential to account for both the dynamic and geometric characteristics of the modeled objects. We leverage the multi-view geometric consistency of 2D Gaussian representations, selecting 2D Gaussians $\mathcal{G}:(\mu, q, s, \sigma, \textit{sh})$ as the foundational primitive for scene representation.

The accuracy of motion is ensured by sparse control points, which are proposed in SC-GS~\cite{huang2024sc}. The sparse controlled points $\mathbf{P}$ and deformation network $\Phi(p_i; \theta)$ are proposed to model the accurate deformation. Where $p_i$ represents the position of the $i$-th control point and $\theta$ denotes the parameters of the deformation network. The control signals at $t_j$ are computed by the MLP $\Phi$:

\begin{equation}
  [R_i^j,T_i^j]= \Phi(p_i,t_j).
\end{equation}

Given $K$ adjacent control points $\mathbf{p} = \{p_{0}, \dots, p_{k}\}$, the 6-DoF transform $\{R^j, T^j\}$ can be computed by a deformation network $\Phi$ applying on $p$, and the deformed 2D Gaussians $\mathcal{G}^j$ at $t_j$ can be interpolated using the LBS~\cite{sumner2007embedded} $\mathcal{I}$ including Eq.~\ref{eq:lbs1} and Eq.~\ref{eq:lbs2} :

\begin{equation}
   \mathcal{G}^j =
   \mathcal{I}\left(\{R^j,T^j\},\mathcal{G} \right).
\end{equation}

This approach ensures that both the temporal dynamics and geometric consistency are preserved, facilitating the extraction of high-fidelity dynamic mesh sequences.

\begin{figure*}[htbp]
\centering
\includegraphics[width=0.9\textwidth]{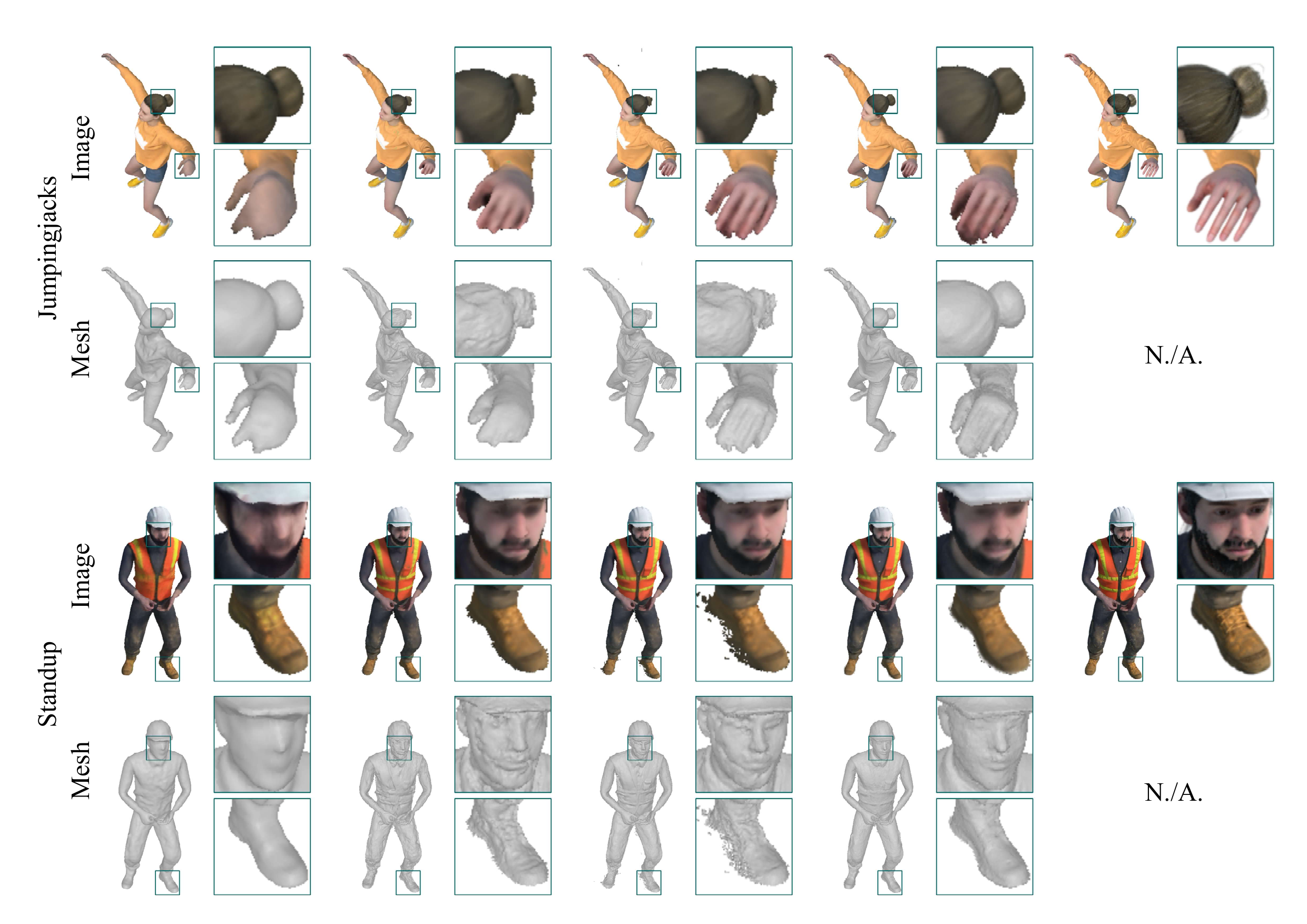}
\includegraphics[width=0.9\textwidth]{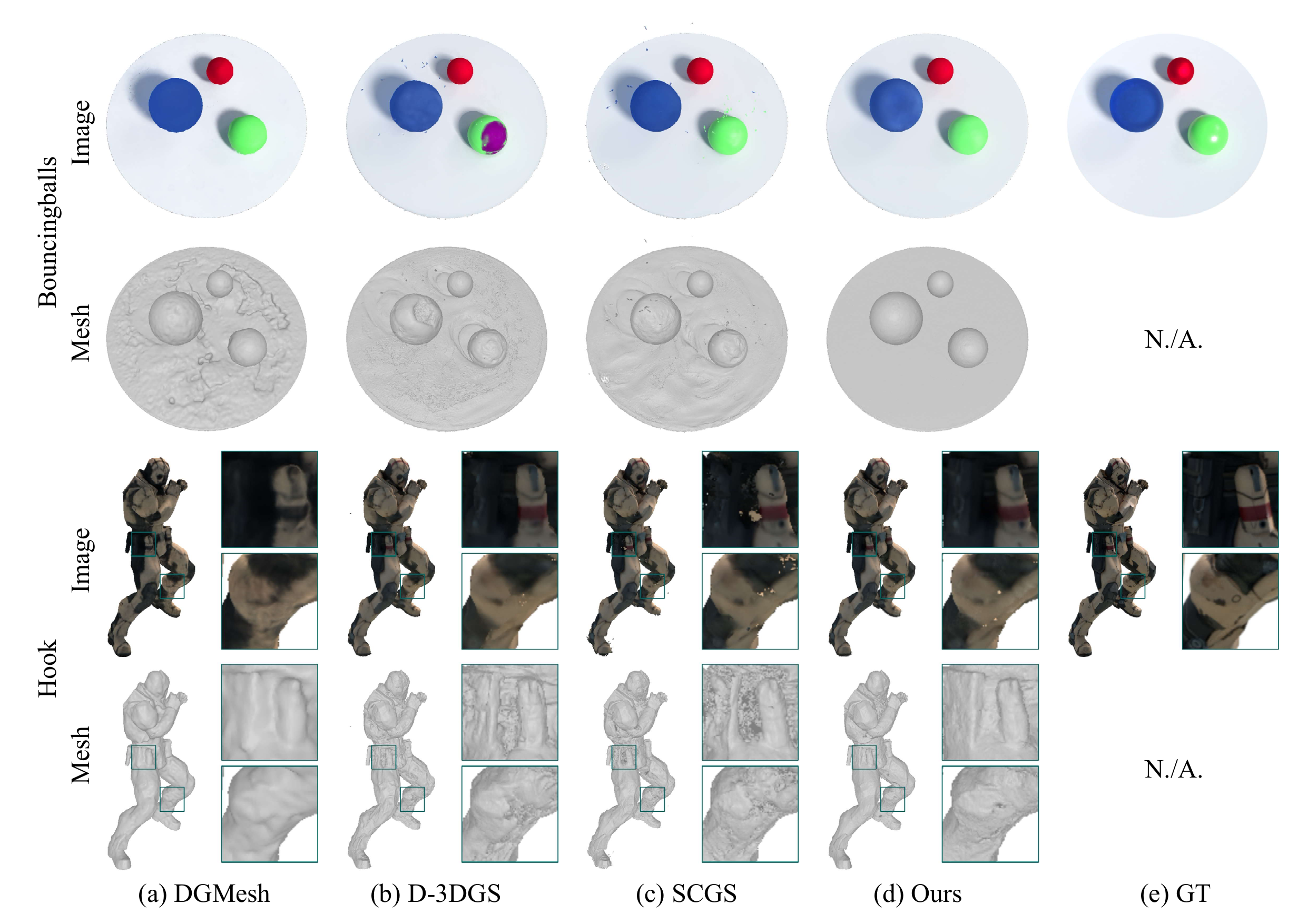}
\vspace{-10pt}
\caption{Mesh and mesh rendering visualization results of the D-NeRF dataset.}
\vspace{-10pt}
\label{fig3}
\end{figure*}

\subsection{Depth filtering}
During the optimization process, \textit{Geometry floaters} manifest as numerous background-colored artifacts, despite the high quality of the rendered image. In the depth image, this issue is evident as noise points located outside the actual object boundaries. Consequently, after mesh extraction, a significant number of floating artifacts appear around the object. To mitigate these artifacts, we propose a masking technique $Flitering$ as mentioned in Eq.~\ref{mask flitering} that involves extracting the object's mask from the rendered high-quality image and applying it to filter the depth image. The process can be formalized as follows:

The extraction of the mask \( \mathbf{M} \) from the rendered image \( \mathbf{I} \) can be mathematically described as follows:

\vspace{-10pt}
\begin{equation}
\mathbf{M}(x) = \mathbf{1}[\mathbf{I}(x) = bg],
\end{equation}
where \( \mathbf{1}[\cdot] \) is the indicator function that returns 1 if the color of \( \mathbf{I}(x) \) is different from the background color \( bg \), and 0 otherwise.

The mask $\mathbf{M}$ is then used to refine the depth image $\mathbf{D}$ by element-wise multiplication:

\vspace{-10pt}
\begin{equation}
    \mathbf{D}' = \mathbf{D} \odot \mathbf{M},
\end{equation}
where $\odot$ represents the element-wise product. The resulting filtered depth image $\mathbf{D}'$ effectively removes the floating noise points, leading to a cleaner mesh extraction with reduced artifacts.

\subsection{Loss Function}
In order to normalize the spatial distribution of Gaussians, we introduce the Depth Distortion and Normal Consistency regularization terms in 2DGS. The Depth Distortion regularization term encourages Gaussians to be distributed at the same depth as much as possible, as shown in the following formula:

\begin{equation}
L_{d}=\sum_{i,j}\omega  _{i}\omega  _{j}\left | z_{i}-z_{j} \right | ,
\end{equation}
where $i$ indexes over intersected splats along the ray, $\omega_{i}$ denotes the blending weight of the intersection point.
$\omega _{i}=\alpha  _{i}\mathcal{G } _{i}(u(x)) {\textstyle \prod_{j=1}^{i-1}}(1-\alpha \mathcal{G }_{j}(u(x))) $ . $z_{i}$ is the depth of the intersection
points.

The Normal Consistency regularization term can make the normals of 2D Gaussians as close as possible to the normals of the object:
\begin{equation}
L_{n}=\sum_{i}\omega _{i}(1-n_{i}^{\mathrm{T}}N),
\end{equation}
where $n_{i}$ represents the normal of Gaussians, $N$ normal
estimated by the nearby depth point $p$, it is calculated by the following formula:

\begin{equation}
N(x,y)=\frac{ \nabla _{x}p \times \nabla _{y}p}{\left | \nabla _{x}p \times \nabla _{y}p \right |}.
\end{equation}

Therefore, the overall loss function is:
\begin{equation}
L_{all}=L_{1}(I_{r},I_{g})+\lambda _{s}L_{ssim}(I_{r},I_{g})+\lambda _{n}L_{n}+\lambda _{d}L_{d},
\end{equation}
where $I_{r}$ represents the rendered image, and $I_{g}$ represents the Ground Truth. $\lambda _{s}$ and $\lambda _{n}$ are the weights of each loss function respectively.

\subsection{Mesh Extraction}

The mesh at any normalized timestamp \( t \in [0, 1] \) can be extracted from the trained model. Given \( t \), the model renders multi-view images \( \mathbf{I}(t) \) and filtered depth images \( \mathbf{D}'(t) \). The mesh is then obtained by applying TSDF~\cite{zhou2018open3d}:

\begin{equation}   
\text{Mesh}(t) = \text{TSDF}\left(\mathbf{I}(t), \mathbf{D}'(t) \right).
\end{equation}

The mesh of an object at any time can be extracted by changing the $t$ value.

\subsection{Post-processing}
Experiments have found that the extracted mesh occasionally has problems with damage and holes, so we use MeshLab to repair small holes in the extracted mesh. The hole-filling algorithm first detects the holes on the mesh, then sets the maximum hole size allowed to be filled, and finally generates new triangular patches to close the holes for the hole boundaries that meet the conditions.

\section{Experiment}

\subsection{Experimental Setups}

We mainly compare some novel view synthesis methods for dynamic scenes, including TiNeuVox-B~\cite{fang2022fast}, DGMesh~\cite{Liu0025a}, DynaSurfGS~\cite{cai2024dynasurfgs}, Deformable 3DGS~\cite{yang2024deformable}, and SCGS~\cite{huang2024sc}. We evaluate the quality of meshes extracted by these methods on the D-NeRF~\cite{pumarola2021dnerf} and DGMesh~\cite{Liu0025a} datasets. Both datasets have large and non-rigid motions and belong to efficient multiview setups as referred to the Dycheck~\cite{dycheck}. Each timestamp has only one view, which is randomly selected. Among them, the D-NeRF dataset does not contain the ground truth of the mesh. Therefore, we report the PSNR, LPIPS, and SSIM of the rendered images by mesh to evaluate the mesh quality in the D-NeRF datasets. For the DGMesh dataset, we report the CD and EMD to evaluate the quality of the generated mesh. It is worth noting that when DGMesh renders an image, two additional MLPs should be maintained when querying the color of the mesh. In addition, DGMesh uses a differentiable Poisson Solver and differentiable Marching Cubes to extract the mesh. Other methods use TSDF to extract mesh like 2D-GS~\cite{huang20242d}.


\subsection{Implementation Details}

We build dynamic 2DGS upon the open-source SCGS~\cite{huang2024sc} codebase. Training iterations for all scenarios are set to 80,000. The training time will take 1 to 2 hours. The number of control points is set to 1024. All experiments are conducted on an NVIDIA RTX A5000 GPU. The hyperparameters $\lambda _{s}$, $\lambda _{n}$ and $\lambda _{d}$ are 1, 0.02 and 1000 respectively. The other hyperparameters are the same as SCGS. When the mesh is extracted, the original dynamic 2DGS representations can be removed. The indicators of TiNeuVox-B and DGMesh come from the DGMesh paper.

\begin{table*}[t!]
    \centering
        \resizebox{\textwidth}{!}{

    \label{tab1}
    \begin{tabular}{cccccccccccccc}
        \toprule
        \multirow{2}{*}{Method}  & \multicolumn{3}{c}{Lego} & \multicolumn{3}{c}{Bouncingballs} & \multicolumn{3}{c}{Jumpingjacks} &  \multicolumn{3}{c}{Hook} \\
          & PSNR$\uparrow$ & SSIM$\uparrow$ & LPIPS$\downarrow$  & PSNR$\uparrow$ & SSIM$\uparrow$ & LPIPS$\downarrow$ & PSNR$\uparrow$ & SSIM$\uparrow$ & LPIPS$\downarrow$ &
          PSNR$\uparrow$ & SSIM$\uparrow$ & LPIPS$\downarrow$\\
        \midrule
        TiNeuVox-B~\cite{fang2022fast} & 21.927 & 0.843 &0.126 &24.819 & 0.947  &0.101& 23.621 & 0.932& 0.075 & 21.429 & 0.908 & 0.085 \\

        D-3DGS~\cite{yang2024deformable} & 22.534 & 0.884&0.115 &23.667 & 0.956  &0.102& 28.878 & 0.972& 0.039 & 27.499 & 0.959 & \cellcolor{yellow}0.048 \\
        SCGS~\cite{huang2024sc}  & \cellcolor{yellow}22.784& \cellcolor{yellow}0.882 &\cellcolor{yellow}0.113& 
        25.231 & 0.956	 &0.103 
        & \cellcolor{yellow}29.534 & \cellcolor{yellow}0.976 &\cellcolor{yellow}0.035 & 27.095 & \cellcolor{yellow}0.961 & 0.048 \\
        
        DGMesh*~\cite{Liu0025a}  & 21.289 & 0.838 & 0.159& 
        \cellcolor{pink}29.145 &\cellcolor{pink}0.972	 &\cellcolor{yellow}0.099&  
        \cellcolor{pink}31.769 & \cellcolor{pink}0.977 &0.045 &\cellcolor{pink}27.884 & 0.954 & 0.074\\

        D-2DGS (Ours)  & \cellcolor{pink}23.293 & \cellcolor{pink}0.887 & \cellcolor{pink}0.112
        &\cellcolor{yellow}27.786 & \cellcolor{yellow}0.969 &\cellcolor{pink}0.073&  29.293 & 0.974&\cellcolor{pink}0.032 & \cellcolor{yellow}27.802 & \cellcolor{pink}0.962 & \cellcolor{pink}0.042 \\
        \midrule
        \multirow{2}{*}{Method}  & \multicolumn{3}{c}{Mutant} & \multicolumn{3}{c}{Standup} & \multicolumn{3}{c}{Trex} & 
        \multicolumn{3}{c}{Hellwarrior} \\
         & PSNR$\uparrow$ & SSIM$\uparrow$ & LPIPS$\downarrow$   & PSNR$\uparrow$ & SSIM$\uparrow$ & LPIPS$\downarrow$   & PSNR$\uparrow$ & SSIM$\uparrow$ & LPIPS$\downarrow$  &
         PSNR$\uparrow$ & SSIM$\uparrow$ & LPIPS$\downarrow$  &\\
        \midrule
        TiNeuVox-B~\cite{fang2022fast} & 22.967 & 0.925 &0.064 &24.263 & 0.941  &0.051& 24.219 & 0.927& 0.070 & 18.657 & 0.917 & 0.118 \\

        D-3DGS~\cite{yang2024deformable}& \cellcolor{yellow}28.342 & 0.957 & 0.048& 
        \cellcolor{yellow}29.946 & \cellcolor{yellow}0.977 	 & \cellcolor{yellow}0.031  &28.061 & 0.958 & 0.054 &\cellcolor{yellow}25.986& \cellcolor{yellow}0.965&0.053 \\
        SCGS~\cite{huang2024sc}& 28.002 & 0.955 &\cellcolor{yellow}0.047&
        29.273 & 0.977	 & 0.032 & 28.521 & \cellcolor{yellow}0.962 & \cellcolor{yellow}0.050  &
        \cellcolor{pink}26.419 & \cellcolor{pink}0.967 	 & \cellcolor{pink}0.048\\

        DGMesh*~\cite{Liu0025a}& \cellcolor{pink}30.400 & \cellcolor{pink}0.968 & 0.055& 
        \cellcolor{pink}30.208 & 0.974	 & 0.051 & \cellcolor{pink}28.951 & 0.959 & 0.065 & 25.460 & 0.959 	 & 0.084  \\

        D-2DGS (Ours)  & 28.120 & \cellcolor{yellow}0.960 &\cellcolor{pink}0.042& 29.512 & \cellcolor{pink}0.977 & \cellcolor{pink}0.028 & \cellcolor{yellow}28.677& \cellcolor{pink}0.967 & \cellcolor{pink}0.043  &25.498 & 0.958 & \cellcolor{yellow}0.049 \\        

        \bottomrule
     
    \end{tabular}
    }
    \caption{Rendering quality of extracted meshes on the D-NeRF~\cite{pumarola2021dnerf} dataset (Background: white). The color of each cell means:  \colorbox{pink}{best},  \colorbox{yellow}{second best}. (*DGMesh uses differentiable Marching Cubes to extract mesh, and the extracted mesh lacks details.)}
    \label{tab:dnef}
    \vspace{-15pt}
\end{table*}

\begin{figure*}
\centering
\includegraphics[width=0.90\textwidth]{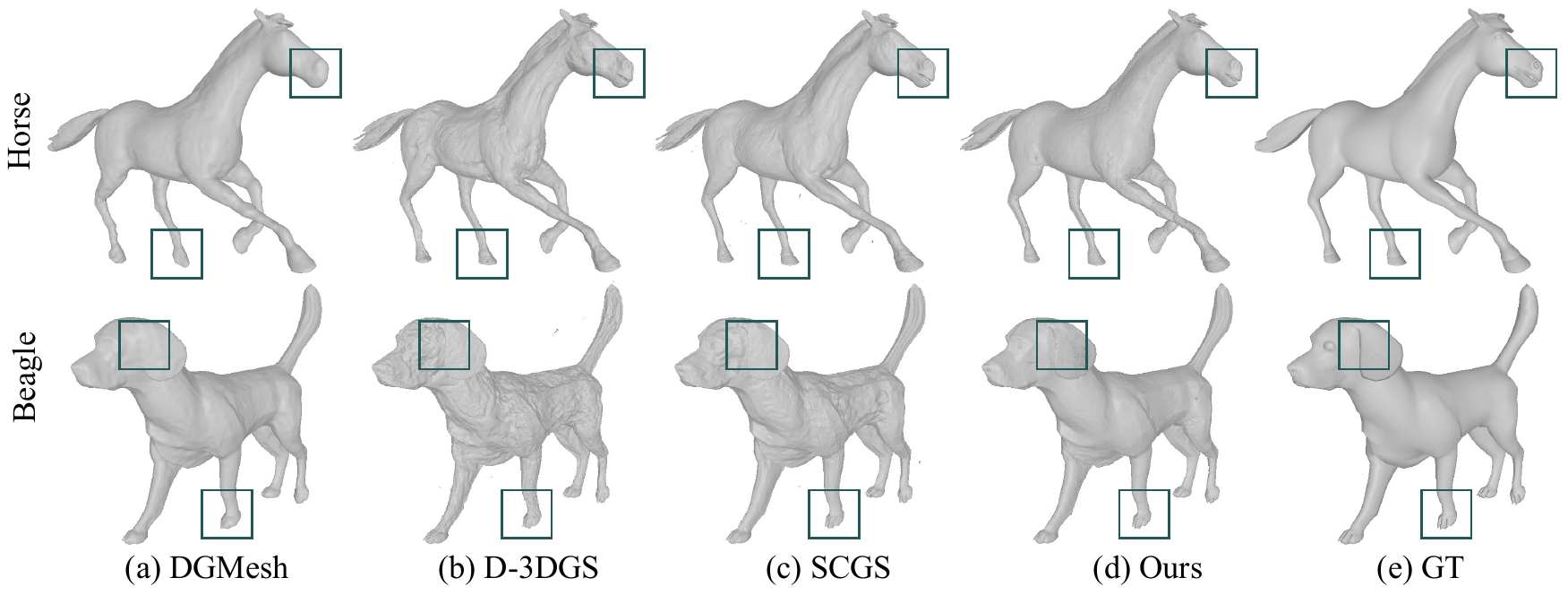}
\vspace{-10pt}
\caption{Mesh visualization results of the DGMesh dataset.}
\label{fig4}
\vspace{-5pt}
\end{figure*}

\begin{table}[]
    \centering
    \resizebox{0.5\textwidth}{!}{
    \begin{tabular}{cccc|ccc}
\toprule
                \multirow{2}{*}{Method}  & \multicolumn{3}{c}{D-NeRF datasets}& \multicolumn{2}{c}{DGMesh datasets}&\\
          & PSNR$\uparrow$ & SSIM$\uparrow$ & LPIPS$\downarrow$ & CD$\downarrow$ & EMD$\downarrow$\\
          \midrule
        TiNeuVox-B~\cite{fang2022fast} & 22.738&0.918 &0.086&1.839&0.130\\
         D-3DGS~\cite{yang2024deformable} & 26.864&0.954 &0.061&0.895&0.115\\
         SCGS~\cite{huang2024sc} &27.107&0.954&0.060&0.856&\cellcolor{yellow}0.115\\
          DGMesh*~\cite{Liu0025a} & 28.137\cellcolor{pink}&0.950\cellcolor{yellow}&0.057\cellcolor{yellow}&\cellcolor{yellow}0.698&0.123 \\
          DynaSurfGS~\cite{cai2024dynasurfgs} &&-&& 0.910&0.123&\\
         D-2DGS(Ours)&\cellcolor{yellow}27.498&\cellcolor{pink}0.957&\cellcolor{pink}0.052&\cellcolor{pink}0.647&\cellcolor{pink}0.112\\
     \bottomrule
    \end{tabular}
    }
    \caption{Average results on both D-NeRF datasets and DGMesh dataset.}
    \vspace{-25pt}
    \label{tab:average_results}
\end{table}

\subsection{Results}
The results on the D-NeRF~\cite{pumarola2021dnerf} dataset are shown in Fig.~\ref{fig3}. The ideal surface should be as smooth as possible while still containing the details of the object.  The mesh extracted by DGMesh~\cite{Liu0025a} has a low resolution and lacks some details of objects, such as the hands of the `jumpingjacks'. Deformable-3DGS and SCGS both use 3D Gaussians as the basic representational primitives, which lack geometric accuracy. The extracted mesh overfits the details and is not geometrically accurate and smooth, such as the heads of the `jumpingjacks'. The three comparative methods are unable to simulate smooth planar and spherical surfaces, as demonstrated by the ``bouncingballs'' example. The sphere appears broken, and the chassis is uneven due to the influence of lighting. In comparison, our Dynamic 2DGS highlights reconstructing rich details and modeling smooth surfaces. 

The rendering results of the mesh are shown in Tab.~\ref{tab:dnef}. When DGMesh renders an image, it needs to query additional MLPs to obtain the color of the mesh, thus achieving higher PSNR. The proposed D-2DGS achieved higher indicators. The average metrics are shown in Tab.~\ref{tab:average_results} and Tab.~\ref{table_model_comparasion}.

As shown in Fig.~\ref{fig4}, on the DGMesh dataset, the proposed method can extract smoother meshes compared to D-3DGS and SCGS. Compared to DGMesh, the proposed method can show the details of objects, such as the mouth of the `horse'. The indicators on the DGMesh dataset are shown in Tab.~\ref{dgmesh_result} and Tab.~\ref{tab:average_results}. The proposed method achieves the lowest CD and EMD value, demonstrating the effectiveness of the proposed method on this dataset.

We evaluate the quality of the mesh from three aspects: whether it is clean (i.e., whether there are floating objects), whether it is smooth, and whether it has details. The image quality rendered by the DGMesh method through Gaussian is relatively low. In addition, the mesh extracted by DGMesh lacks details. The mesh extracted by D-3DGS and SCGS is prone to floaters. In addition, its surface is uneven, prone to potholes, and not smooth enough. In comparison, the mesh extracted by our method not only has better details but is also relatively smooth.

\vspace{-5pt}

\begin{table}[h!]
    \centering
   
    \label{tab1}
    \resizebox{0.5\textwidth}{!}{
    \begin{tabular}{cccccccc}
        \toprule
        \multirow{2}{*}{Method}  & \multicolumn{2}{c}{Duck} & \multicolumn{2}{c}{Horse} & \multicolumn{2}{c}{Bird} & \\
          & CD$\downarrow$ & EMD$\downarrow$  & CD$\downarrow$ & EMD$\downarrow$   & CD$\downarrow$ & EMD$\downarrow$ &  \\
        \midrule
        TiNeuVox-B~\cite{fang2022fast} & 0.969 &0.059 &1.918 &  0.246 &8.264 &0.215  \\
        D-3DGS~\cite{yang2024deformable} & 1.064 &0.076 &\cellcolor{yellow}0.298 &  \cellcolor{pink}0.114 &0.427 &0.123  \\
        SCGS~\cite{huang2024sc}   &  1.001&0.074  & \cellcolor{pink}0.232&\cellcolor{yellow}0.118  &\cellcolor{yellow}0.336 &\cellcolor{yellow}0.121  \\
        DGMesh~\cite{Liu0025a}  & \cellcolor{pink}0.782 & \cellcolor{pink}0.048 & 0.299 & 0.168	  &  0.557 & 0.128  \\
        D-2DGS (Ours) &\cellcolor{yellow}0.841 & \cellcolor{yellow}0.057  &0.327& 	0.134 &  \cellcolor{pink}0.328 & \cellcolor{pink}0.110  \\
        \midrule
        \multirow{2}{*}{Method}  & \multicolumn{2}{c}{Beagle} & \multicolumn{2}{c}{Torus2sphere} & \multicolumn{2}{c}{Girlwalk} & \\
         & CD$\downarrow$ & EMD$\downarrow$   & CD$\downarrow$ & EMD$\downarrow$  & CD$\downarrow$ & EMD$\downarrow$  &\\
        \midrule
        TiNeuVox-B~\cite{fang2022fast} & 0.874 &0.129 &2.115 &  0.203 &0.568 &0.184 \\
        D-3DGS~\cite{yang2024deformable}& 0.585 & \cellcolor{yellow}0.101  & 2.655 & 0.155 & 0.340  & \cellcolor{yellow}0.122 \\
        SCGS~\cite{huang2024sc} & \cellcolor{pink}0.528 &\cellcolor{pink}0.100   &2.716 &0.155    &\cellcolor{pink}0.321& 0.124   \\
        DGMesh~\cite{Liu0025a}& 0.626 & 0.114 & \cellcolor{yellow}1.515 &\cellcolor{pink}0.126 	  & 0.406 & 0.153    \\
        D-2DGS (Ours)  & \cellcolor{yellow}0.553 & 0.115  &\cellcolor{pink}1.502  & \cellcolor{yellow}0.139 &\cellcolor{yellow}0.332& \cellcolor{pink}0.116      \\
        \bottomrule
     
    \end{tabular}
    }
    \caption{Mesh reconstruction quality comparison results of DGMesh dataset.}
    \label{dgmesh_result}
    \vspace{-20pt}
\end{table}




\begin{table}
    \centering
 \resizebox{0.45\textwidth}{!}{
  \begin{tabular}{@{}cccccc@{}}
    \toprule
     Method &Train Time&Render Speed&Memory&Storage&\\
    \midrule
    TiNeuVox-B~\cite{fang2022fast}  & 28 mins & 1.5 FPS & 5574 MB& 48 MB\\
    D-3DGS~\cite{yang2024deformable}  & 10 mins &190 FPS  &4270 MB&10 MB\\
    SCGS~\cite{huang2024sc} &16 mins& 215 FPS&3200 MB&26 MB\\
    DGMesh~\cite{Liu0025a}  & 90 mins & 140 FPS&9425 MB &17 MB\\
    D-2DGS (Ours) & 	32 mins & 71 FPS &3690 MB&20 MB\\
    \bottomrule
  \end{tabular}
  }
  \caption{Comprehensive comparison of different methods. }
 \vspace{-15pt}
 \label{table_model_comparasion}
\end{table}

\begin{figure}
\centering
\includegraphics[width=0.49\textwidth]{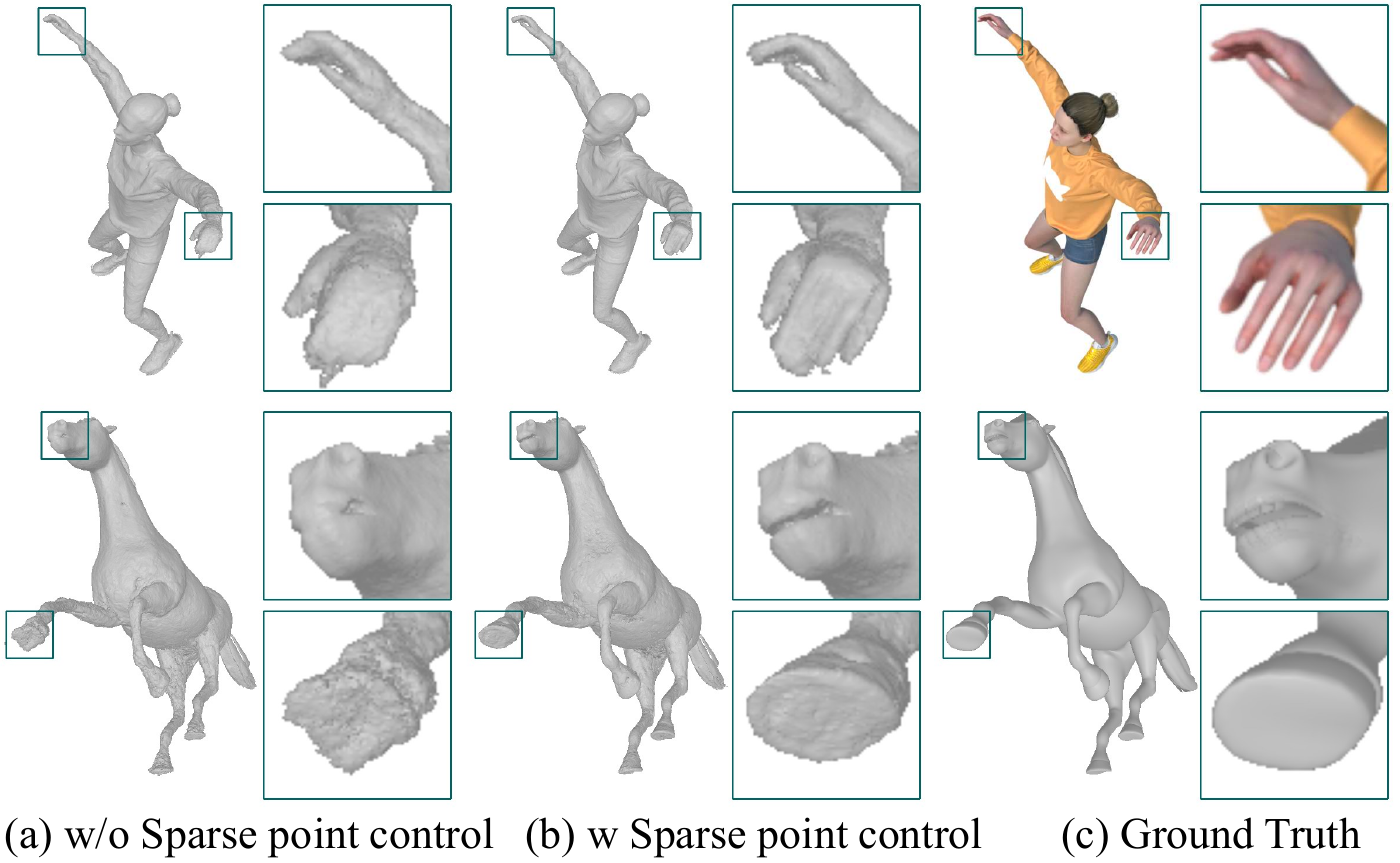}
\caption{Ablation Study on Sparse point control.}
\label{fig:ablation1}
\vspace{-5pt}
\end{figure}

\subsection{Ablation Study}

\paragraph{Sparse point control.} We conducted ablation experiments on sparse point control, where we removed the sparse point control and directly modeled the offset for 2DGS using an MLP as the offset field. As shown in Fig.~\ref{fig:ablation1}, in areas with significant motion, such as a person's hand and a horse's hoof, sparse point control allows for the extraction of a more detailed mesh. This shows that sparse point control can model finer motions, thus helping to extract more detailed meshes. 

\begin{figure}
\centering
\includegraphics[width=0.47\textwidth]{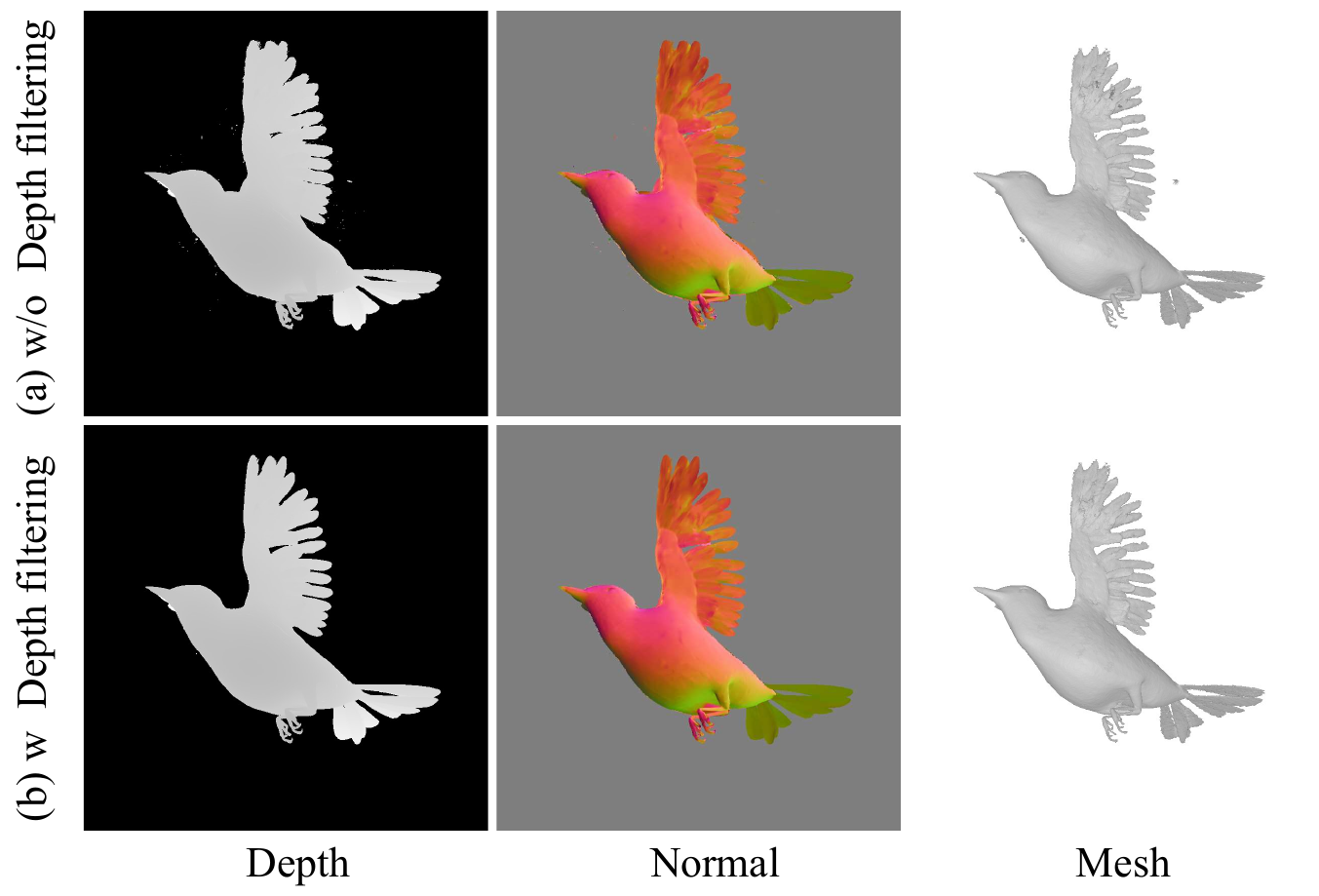}
\vspace{-10pt}
\caption{Ablation Study on Depth filtering.}
\label{fig6}
\vspace{-10pt}
\end{figure}

\paragraph{Depth filtering.} We removed the mask on the Depth image and then conducted experiments. As can be seen in Fig.~\ref{fig6} (a), there are many noise points around the object in the rendered depth image, which may be caused by the offset field offsetting the 2D Gaussians. After extracting the mesh through TSDF fusion, some floating objects will appear around the mesh of the object. With the help of high-quality renderings, the object mask is extracted, and the depth image is masked. The noise in the depth image can be filtered out, thereby removing the floaters in the mesh, as shown in Fig.~\ref{fig6} (b). Experiments show that the proposed depth filtering method is beneficial for removing floaters and obtaining a cleaner mesh.

\begin{figure}
\centering
\includegraphics[width=0.47\textwidth]{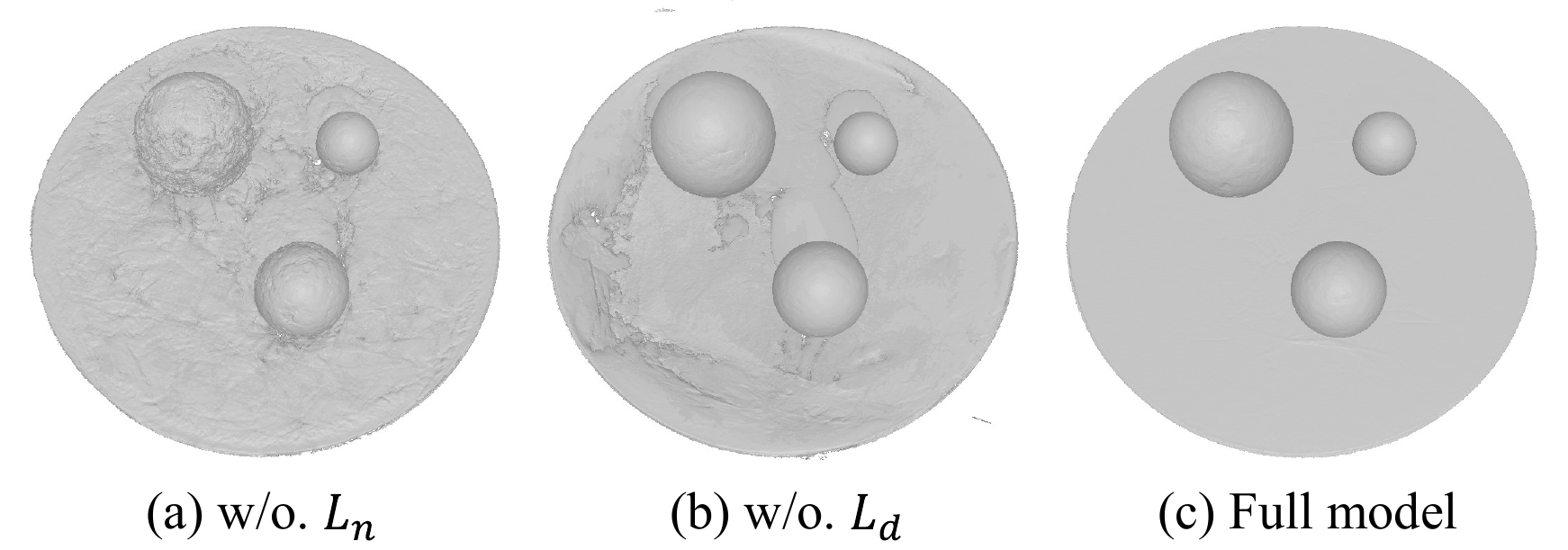}
\vspace{-5pt}
\caption{Ablation Study on loss function.}
\label{fig:loss}
\vspace{-10pt}
\end{figure}

\paragraph{Loss function.} We conduct ablation experiments on various loss functions to verify the effectiveness of the loss functions. As shown in Fig.~\ref{fig:loss}, the two regularization terms in the loss function are important for extracting high-quality meshes. Among them, the normal consistency regularization term can make the normal of the Gaussian sphere more accurate, thereby making the surface of the mesh smoother. The Depth Distortion term can make the Gaussians distributed on the plane, thus reducing the surface damage. The corresponding metrics are shown in Tab.~\ref{tab:ablation_study}. It can be found from the table that adding two regularization terms to the loss function significantly reduces the CD value and improves the quality of the mesh.

\vspace{-5pt}

\begin{figure}
\centering
\includegraphics[width=0.48\textwidth]{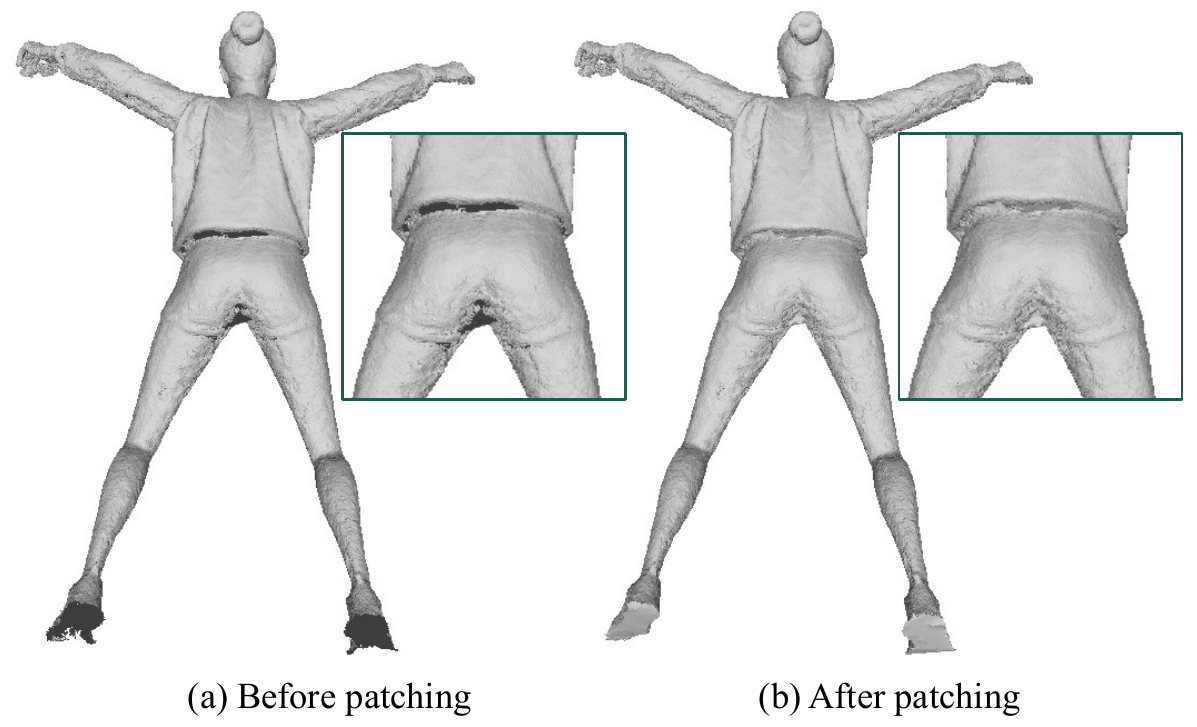}
\vspace{-10pt}
\caption{Repair holes in the mesh.}
\label{fig:patching}
\vspace{-5pt}
\end{figure}

\paragraph{Post-processing.} Since the input view of the model is sparse, the extracted mesh is prone to holes in some areas that are difficult to see. Therefore, we use some post-processing methods to repair the holes that may appear in the mesh to improve the integrity of the mesh. As shown in Fig.~\ref{fig:patching}, we used MeshLab to repair the holes in the mesh. As can be seen from the figure, simple post-processing operations can fill in the holes that may appear in the reconstructed mesh.

\begin{table}[]
    \centering
    \resizebox{0.45\textwidth}{!}{
    \begin{tabular}{cccccc}
\toprule
                \multirow{1}{*}{Method}    & CD$\downarrow$ & EMD$\downarrow$ &  {Method}    & CD$\downarrow$ & EMD$\downarrow$ \\
          \midrule
         w/o DF &0.425 &0.110&w/o DD&0.350&0.109 \\
         w/o NC &0.443&0.101& Full &0.328 &0.110\\
     \bottomrule
    \end{tabular}
    }
    \caption{Ablation experiment results of the “bird” sample in the DGMesh dataset. (DF: Depth filtering, NC: Normal Consistency, DD: Depth Distortion)}
      \vspace{-20pt}
    \label{tab:ablation_study}
\end{table}



\section{Limitations and Future work}

Though dynamic 2D Gaussians can model the accurate dynamic surface, there are several limitations: (1) Some holes exist in the extracted mesh. We need some post-processing operations to repair these holes. Future work will pursue direct hole-free generation to bypass repair entirely. (2) Under ultra-sparse (e.g., monocular) views, robust priors and low-rank motion can enhance accuracy. Recently, 3D foundation models~\cite{xiang2025structured,tochilkin2024triposr,li2025triposg,ye2025hi3dgen} show the great potential to generate high quality mesh from a static image. We believe that reconstruct high quality mesh from the dynamic multiview input will be coming soon.

\section{Conclusion}

In this paper, we propose Dynamic 2D Gaussians to extract high-quality meshes from dynamic 2D images input. We use 2D Gaussians as the basic representation primitives and use sparse control points to control 2D Gaussians to model the motion of objects. We extract masks from high-quality rendering data and use these masks to filter out noise in the depth map. Finally, we use TSDF to extract the mesh of the object at any time. Experiments show that the mesh extracted by the proposed method has high details and a smoother surface.

\begin{acks}
This work was supported by National Natural Science Foundation of
China (No. 62376102). 
\end{acks}
\vspace{-5pt}

\bibliographystyle{ACM-Reference-Format}
\bibliography{sample-base}

\clearpage

\section{Appendix}

In the supplementary material, we mainly provide additional experimental results in Sec. 1. Then more discussions are conducted in Sec. 2.  \textbf{In addition, we provide some visual comparison videos in the support materials folder.}

\subsection{Additional Experimental Results}

In the D-NeRF dataset, the metrics of the RGB images rendered from the extracted meshes do not fully reflect the quality of the meshes. Therefore, we provide additional visualization results, as shown in Fig.~\ref{result}. DGMesh exhibits relatively poorer performance in modeling object details. The meshes extracted by Deformable-3DGS and SCGS exhibit numerous pits and holes on the surface. In comparison, the meshes extracted by the proposed Dynamic 2D Gaussians approach capture object details while maintaining a smoother surface, reflecting better geometric properties.

\begin{figure}[htbp]
\centering
\includegraphics[width=0.5\textwidth]{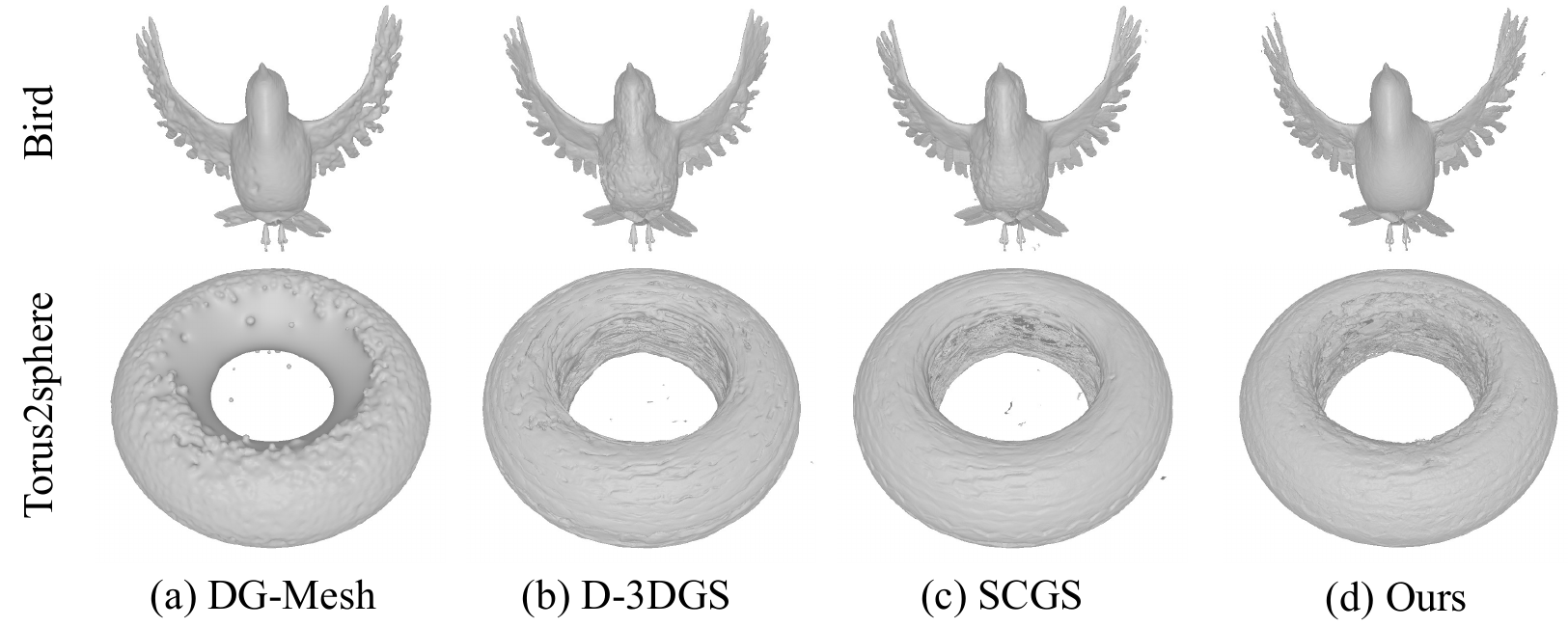}
\vspace{-10pt}
\caption{Visualization results of DGMesh dataset.}
\label{fig:DGmeshimage}
\vspace{-10pt}
\end{figure}

\begin{figure}[h]
\centering
\includegraphics[width=0.46\textwidth]{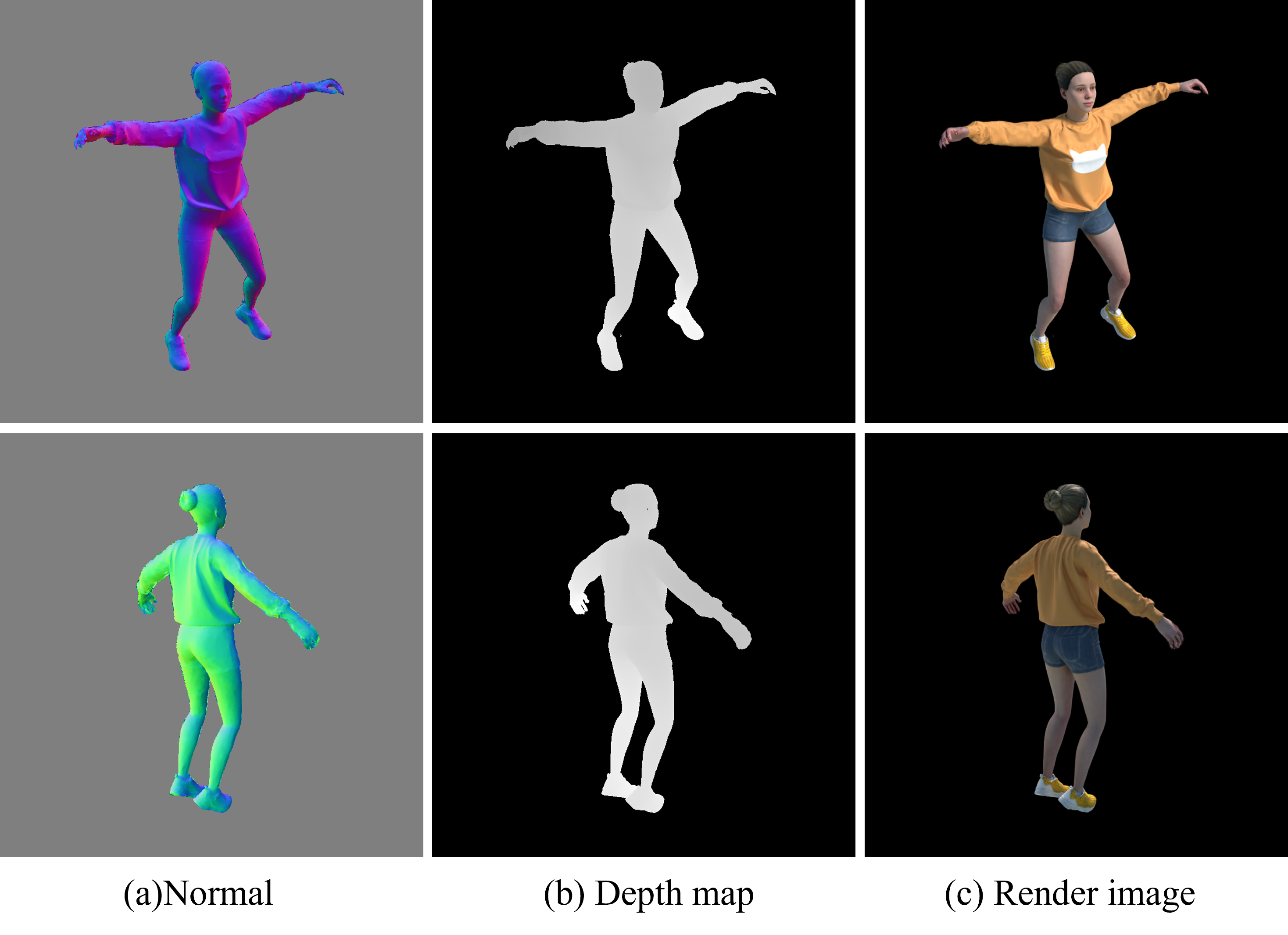}
\caption{Rendered images, normals, and depth maps.}
\label{fig:render_image}
\vspace{-7pt}
\end{figure}

\begin{figure}[htbp]
\centering
\includegraphics[width=0.46\textwidth]{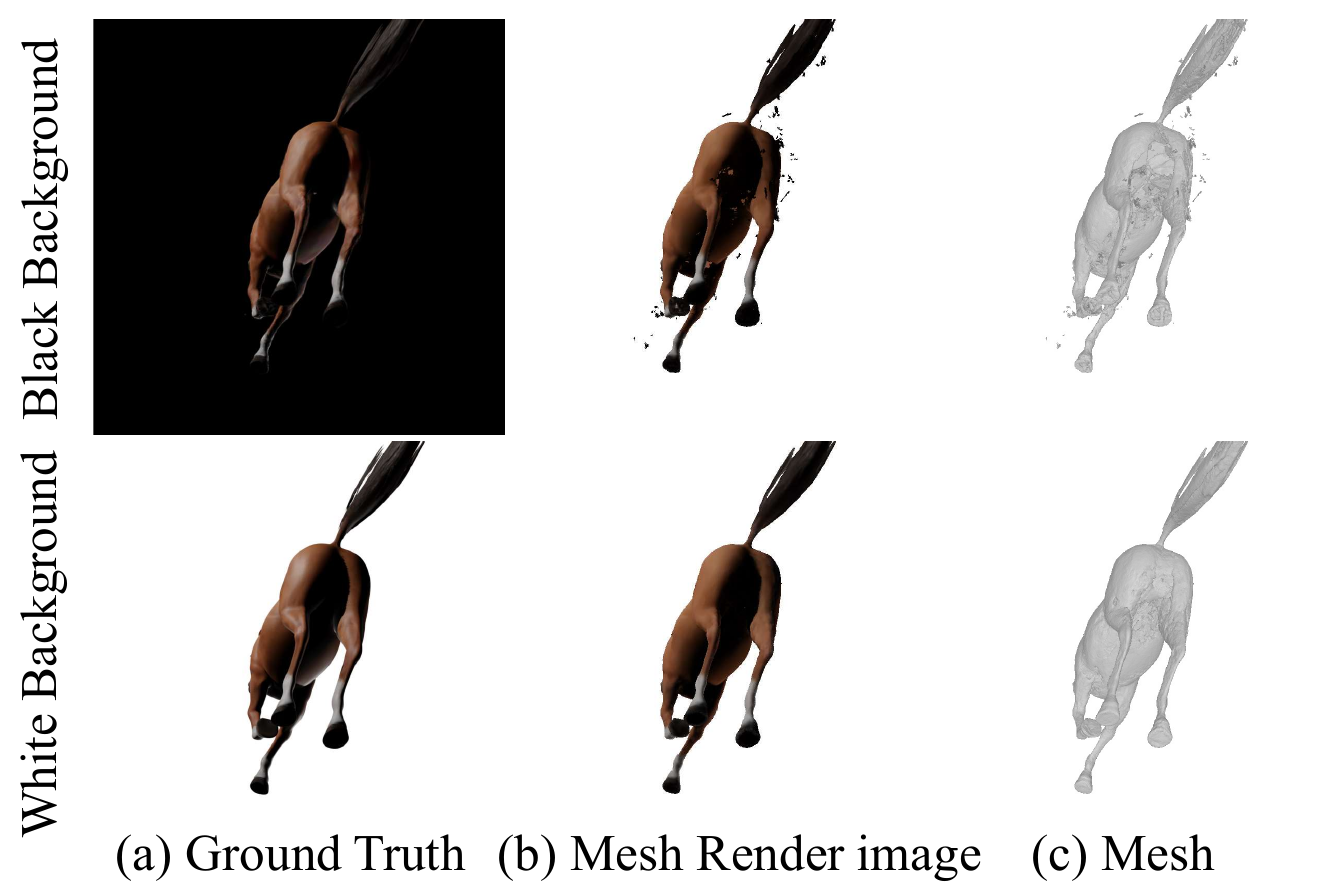}
\caption{Comparison of results with different background colors.}
\label{background}
\end{figure}

\begin{figure}[htbp]
\centering
\includegraphics[width=0.48\textwidth]{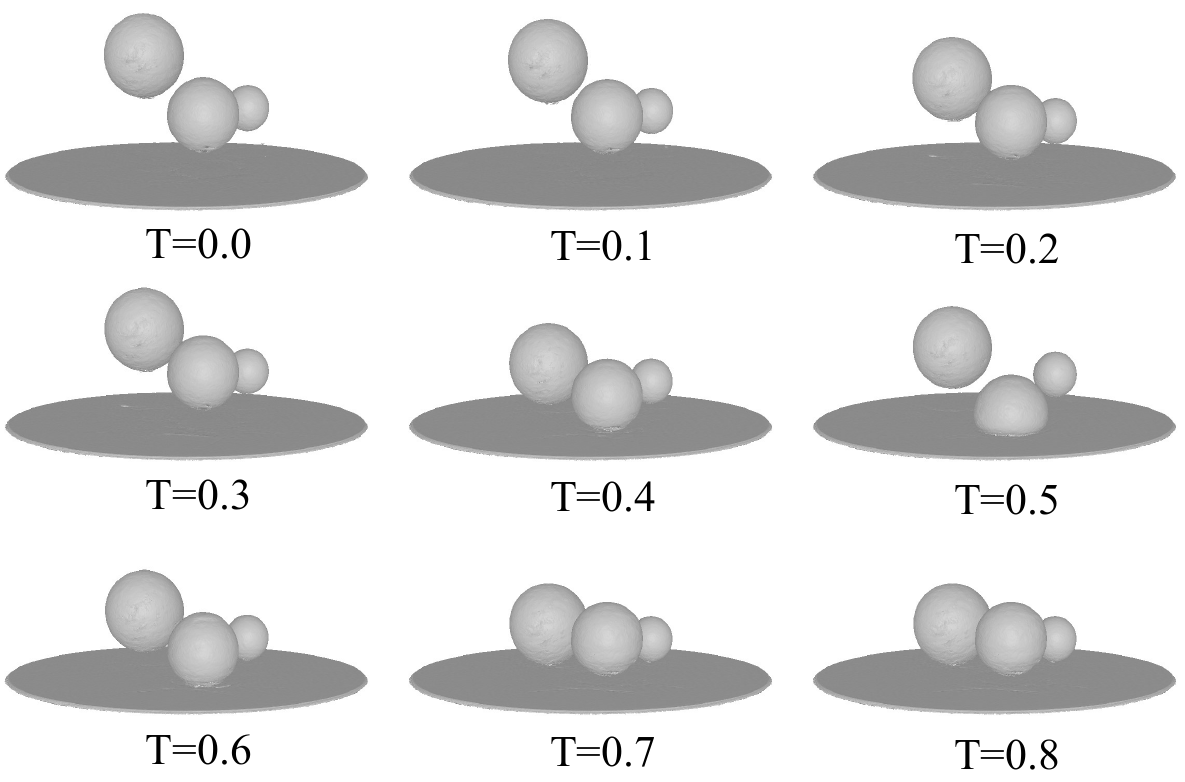}
\caption{Time-Aware Mesh.}
\vspace{-15pt}
\label{time-mesh}
\end{figure}

\begin{figure*}[htbp]
\centering
\includegraphics[width=0.90\textwidth]{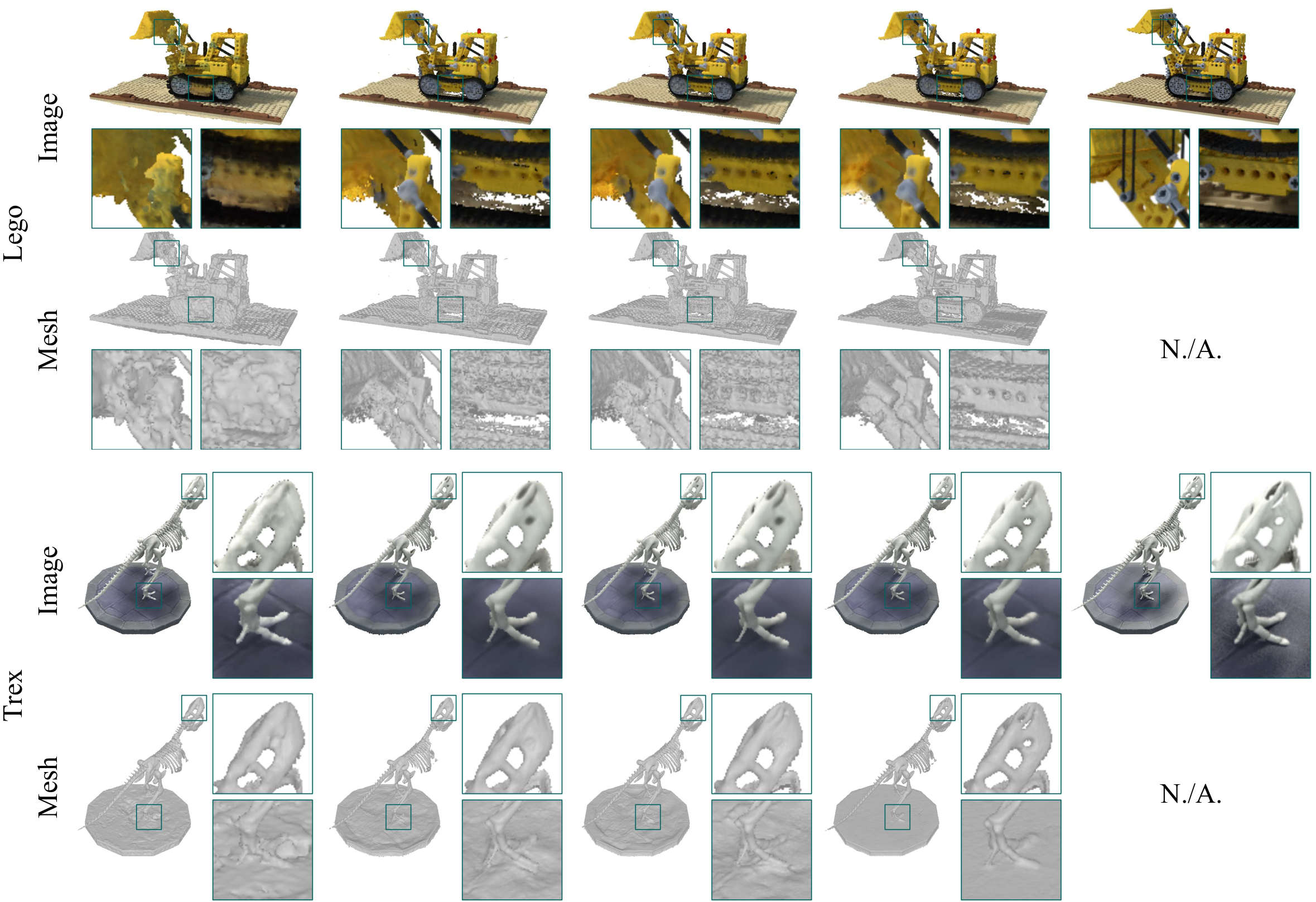}
\includegraphics[width=0.90\textwidth]{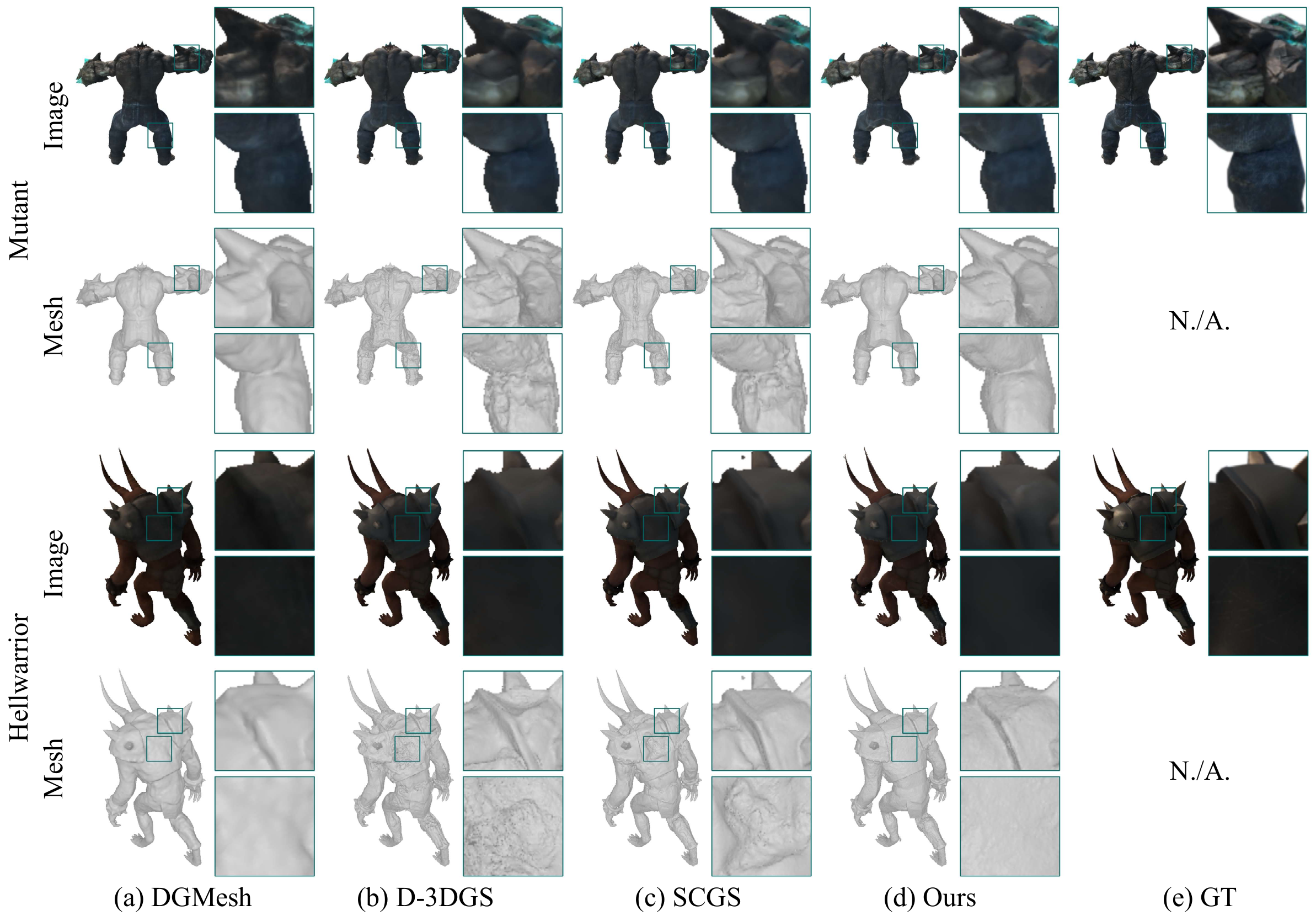}
\caption{Mesh and mesh rendering visualization results of the D-NeRF dataset.}
\label{result}
\end{figure*}

Since the data of the DGMesh dataset is relatively simple and the surface is smooth, it is difficult to reflect the advantages of the proposed method in terms of metrics. The visual quality shows that our extracted surfaces appear smoother than SCGS. The DGMesh method shows advantages in the Duck and Torus2sphere examples, but performs relatively poorly on the other cases. Although DGMesh has a high metrics in Torus2sphere, from the visualization results in Fig.~\ref{fig:DGmeshimage}, DGMesh does not show a particularly good reconstruction effect in the Torus2sphere example.

The proposed method can render Gaussians to obtain normal depth maps and high-quality RGB images of dynamic objects, as shown in Fig.~\ref{fig:render_image}.

\subsection{Discussion and Application}

\paragraph{Background Color.} Training models with different background colors can have some influence on the extraction of the mesh. The background color should be selected to have significant contrast with the color of the object itself. For the training images, the background color should be selected to have significant contrast with the color of the object itself. As shown in Fig.~\ref{background}, if a black background is used, the shadows on the object are difficult to distinguish from the background, making it challenging to extract the object's mask from the rendered images. This then makes it difficult to filter out floating artifacts in the depth maps, ultimately leading to the presence of many floating artifacts in the extracted meshes.


\paragraph{Time-Aware Mesh.} Our method supports the extraction of dynamic mesh sequences, as shown in Fig.~\ref{time-mesh}. The extracted precise temporal geometry provides the potential for analyzing the motion characteristics and dynamics of objects.

\end{document}